\documentclass[10pt,letterpaper]{article}
\usepackage{color}

\usepackage{cogsci}
\usepackage{pslatex}
\usepackage{apacite}
\usepackage{amsmath,amsfonts,amssymb}

\usepackage{marvosym}
\usepackage{graphicx}
\usepackage{epsf}

\title{Counterfactuals, indicative conditionals, and negation under uncertainty:\\ Are there cross-cultural differences?}
 
\author{{\large \bf Niki Pfeifer (niki.pfeifer@lmu.de)} \\
Munich Center for Mathematical Philosophy, LMU Munich, Germany
  \AND {\large \bf Hiroshi Yama (yama@lit.osaka-cu.ac.jp)} \\
   Graduate School of Literature and Human Behavioral Sciences, Osaka City University, Japan}

\begin{document}

\maketitle

\begin{abstract}
In this paper we study selected argument forms involving
counterfactuals and
indicative conditionals
under uncertainty. We selected argument forms
to explore whether people with an Eastern cultural background  reason
differently about conditionals compared to Westerners,  because of the differences in the location of negations. In a $2\times 2$ between-participants design, 63 Japanese
university students were allocated to four groups, crossing
indicative conditionals and counterfactuals, and each presented  in
two random task orders.
 The data show close agreement between the responses of 
Easterners and Westerners. The modal responses provide strong
support for the hypothesis that conditional probability is the best
predictor for counterfactuals and indicative conditionals. Finally,
the grand majority of the responses are probabilistically coherent,
which endorses the psychological plausibility of choosing \emph{coherence-based
probability logic} as a  rationality framework for psychological
reasoning research.


\textbf{Keywords:} 
argument forms; cross-cultural comparison; counterfactuals; indicative conditionals; negation; probability logic;
reasoning under uncertainty
\end{abstract}

\section{Introduction}

In this paper we study selected argument forms involving
counterfactuals and
indicative conditionals
under uncertainty. The aim is to explore potential cross-cultural
differences in human reasoning about conditionals and negation under
uncertainty  between Easterners and Westerners. So far, cross-cultural
differences in reasoning involving negations have been described in   the classical-logic based (old) paradigm psychology of reasoning literature \cite<see, e.g.,>{nisbett01,
norenzayan02,peng99,yamaIP}. These previous studies  demonstrate that Westerners are inclined to engage in rule-based reasoning whereas Easterners are apt to engage in intuitive or dialectical reasoning. In other words, Easterners are more likely to consider contradictory premises dialectically than Westerners. However, \citeA{zhang15} report that Easterners are not actually more dialectical when they meet contradictory opinions, but they believe due to cultural reasons that dialectical thinking is wiser than Westerners. Because  contradictory premises are not used in this experiment, we do not make   predictions concerning  whether Easterners reason more  dialectical or not  \cite<see, e.g.,>{peng99}. Rather, we explore whether the location of negation in the context of conditionals impacts on  reasoning and whether our Japanese sample differs from corresponding data of Western samples. Moreover, if Japanese people see a stronger cultural value in dialectical thinking, it is plausible to assume that they may hesitate to show stronger confidence in the correctness of their judgments. Our study
presents one of the first attempts \cite<see also>{yamaIP} to
identify cross-cultural differences within the framework of  the new probability-based paradigm psychology
of reasoning.   

Among the various ways of expressing
and using counterfactuals \cite<see, e.g.>{declerck01}, we restrict our investigation of
counterfactuals to conditionals in subjunctive
mood, where the grammatical structure implies that the counterfactual's  \emph{antecedent} ($A$) is factually
false. For instance, consider the utterance of the following counterfactual in the
context of a randomly drawn poker card: 
\begin{equation}
\begin{split} \label{CounterfEx}
\text{If the drawn 
  card were to show an ace } (A), \\ \text{then it would show
  spades }
  (C)\, .
\end{split}
\end{equation}
The grammatical structure of  (\ref{CounterfEx}) pragmatically entails that the drawn card is not an ace
($\neg A$), i.e., the  antecedent $A$ of (\ref{CounterfEx})  is false.  By  ``indicative conditional'' we mean an ``if--then'' statement of the form
\emph{If $A$, then $C$}, e.g.,
\begin{equation}
\begin{split} \label{IndicativeEx}
\text{If the drawn
  card shows an ace, then it shows
  spades}\, .
\end{split}
\end{equation} Contrary to the counterfactual~(\ref{CounterfEx}), the
indicative conditional~(\ref{IndicativeEx}) does not imply whether
the card actually shows an ace or not.
While the core meaning of indicative conditionals was equated
with the semantics of the \emph{material conditional} in the classical
logic-based 
paradigm (or ``old'') psychology of reasoning \cite<see,
e.g.,>{braine98,johnsonlaird83,rips94,wason72}, our work is located in the new paradigm psychology of reasoning, where conditionals are
interpreted as \emph{conditional probability} assertions \cite<see,
e.g., >{elqayam16,oaksford07,over09,pfeifer13b}. Instead of using 
(fragments of) classical logic, the new paradigm psychology of
reasoning uses probability theory as a rationality
framework. Probability as a rationality framework is psychologically
and philosophically
appealing for many reasons \cite<see, e.g.,>{pfeifer14}. Let us
mention three of them. 

First, probability theory  allows for managing
\emph{degrees of belief}  instead of restricting belief to the two values \emph{true} and \emph{false} as
in the case of bivalent classical logic. Thus, probability
theory provides a much richer framework to study
conditionals. 
It allows for analysing different psychological predictions concerning
conditionals: not only in terms of the material conditional  ($A\supset C$) and the conjunction
($A\wedge C$) as defined in classical logic, but also in terms of the conditional
event ($C|A$), as defined in coherence-based probability logic \cite<see,
e.g.,>{coletti02,gilio16,pfeifer09b}. Table~\ref{FIG:ttables} presents
the truth conditions of these three interpretations. 
Note that the
conditional event cannot be expressed in classical bivalent
logic. We hypothesise that the degree of belief in a conditional \emph{If $A$, then $C$} is
interpreted by a suitable conditional
probability assertion ($p(C|A)$) and neither as the probability of the
material conditional ($p(A\supset C)$) nor as the probability of the
conjunction ($p(A \wedge C)$).  We will test these three
interpretations in the following experiment.  
\begin{table}[!ht]
\begin{center} 
\caption{Truth tables for the material conditional $A \supset C$ interpretation,
 the conjunction $\wedge$ interpretation and the conditional event
 interpretation $C|A$ of a (counterfactual) conditional \emph{If $A$ (were the case), then
   $C$ (would be the case)}.} 
\label{FIG:ttables} 
\vskip 0.12in
\begin{tabular}{ccccc}\hline
$A$ & $C$ & $A \supset C$& $\wedge$& $C|A$\\\hline
true &true &true &true &true  \\
true &false &false &false &false  \\
false &true &true &false & undetermined  \\
false &false &true &false &undetermined  \\\hline
\end{tabular}
\end{center}
\end{table}

Second, probability logic blocks so-called paradoxes of the material
conditional \cite<see, e.g.,>{pfeifer13}. For example, $\neg A$ (``not-$A$'') logically entails $A
\supset C$. It is easy to imagine natural language instantiations for
$A$ and $C$, where this inference  appears counterintuitive. The
paradox arises, when the material conditional is used to formalize a
natural language conditional. In probability logic, the inference from
$p(\neg A)=x$ to $p(C|A)$ is probabilistically non-informative, i.e.,
if $p(\neg A)=x$, then $0\leq p(C|A)\leq 1$ is coherent; hence, the
paradox is blocked
\cite{pfeifer13}. This  also matches experimental data based on samples
involving Westerners
\cite{pfeifer10b,pfeifertulkki17}. Note that the paradox is not blocked if the
conditional probability (conclusion) is replaced by
$p(A\supset C)$ or by $p(A \wedge C)$. A subgoal of this paper is to explore  how Japanese
participants reason about this paradox.

Third, probability allows 
for retracting conclusions in the light of new evidence while
classical logic is monotonic (i.e., adding a premise to a logically
valid argument can only increase the set of conclusions). The
suppression effect \cite<see, e.g.,>{byrne89,stenning05} 
illustrates peoples' capacity to retract conclusions if new premises
are learned. Moreover, experimental data suggests that most people
satisfy 
basic nonmonotonic reasoning postulates of System~P \cite<see,
e.g.>{benferhat05,pfeifer,pfeifer10a}. The rules of  System~P describe
formally basic principles any system of nonmonotonic reasoning
should satisfy \cite{kraus90} and different semantics were developed,
including probabilistic ones. Probabilistic semantics  postulate that
conditionals should be represented by conditional probability assertions \cite<see,
e.g.,>{adams75,gilio02}. Interestingly, inference rules which are
(in)valid in System~P are also (in)valid in standard systems of
counterfactual conditionals \cite<like>{lewis73}. This convergence
shows a close relation between conditional probabilities and
counterfactuals. Compared to the big number of psychological investigations on indicative conditionals
\cite<for overviews see, e.g.,>{evans04,Nickerson2015}, studies
on adult reasoning about counterfactuals are surprisingly
rare \cite{over07b,pfeifer15a,pfeifertulkki17}. Our study sheds  new
light by adding a cross-cultural perspective on indicative conditionals and counterfactuals. 

\begin{table}[!ht]
\begin{center} 
\caption{Task names, their abbreviations and formal structures used in
  the experiment, where $\neg$ denotes negation, $\rightarrow$ is a
  placeholder for denoting the
  indicative conditional or the counterfactual, $\supset$ denotes the material
conditional, $\therefore$ denotes ``Therefore''.} 
\label{TAB:tasks} 
\vskip 0.12in
\begin{tabular}{ll}\hline
Task name (abbreviation) &\!\!\!Argument form \\\hline
Aristotle's thesis \#1 (AT1)     &\!\!\!it's not the case that:$(\neg A \rightarrow A)$\!\!\!           \\
Aristotle's thesis \#2 (AT2)     &\!\!\!it's not the case that:$(A \rightarrow \neg A)$\!\!\!    \\
Negated Reflexivity  (NR)  &\!\!\!it's not the case that:$(A \rightarrow A)$           \\
From ``Every''  to  ``If'' (EIn) &\!\!\!Every $S$ is $P$ $\therefore$ $S\rightarrow \neg P$  \\ 
From  ``Every''  to ``If'' (EI) &\!\!\!Every $S$ is $P$ $\therefore$ $S\rightarrow P$   \\
Modus Ponens  (MP)&\!\!\!$A$, $A\rightarrow C$ $\therefore$ $C$  \\
Negated MP (NMP)    &\!\!\!$A$, $A\rightarrow C$ $\therefore$ $\neg C$   \\
Paradox (Prdx)  &\!\!\!$\neg A$ $\therefore$ $A \rightarrow C$    \\\hline
\end{tabular}
\end{center}
\end{table}

Table~\ref{TAB:tasks} lists the task names, their abbreviations,  and their underlying logical
form used in our experiment. All
argument forms were investigated previously in the literature on Western
samples. Each argument form is suitable for indicative and subjunctive formulations.   They are carefully selected to distinguish between the
material conditional, conjunction and conditional event interpretation
of conditionals. Tasks AT1, AT2, and NR \cite<adapted from>{pfeifer12x} are about negating
conditionals. Note that  there are
two ways to negate material conditionals, namely the wide scope
negation of material conditionals
(i.e., $A \supset C$ can be negated by $\neg (A \supset C)$) and  the
narrow scope negation of material conditionals (i.e., $A \supset C$ is negated
by negating its consequent $C$:  $A \supset \neg
C$). Table~\ref{TAB:percentages} lists the normative predictions of the
different argument forms. Averaging the percentages of
responses in three studies reveals that 73\% of the participants in
task AT1,  75\% in
task AT2,  and 80\% of the participants in
task NR   responded probabilistically coherently according to   the
conditional probability interpretation \cite{pfeifer12x,pfeifer15a,pfeifertulkki17}.

Task EI (resp., task EIn) connects the basic syllogistic sentence type ``\emph{Every
$S$ is $P$}'' with associated conditionals (resp., conditionals involving negations) in the indicative and in the
counterfactual form. The motivation for these tasks is to shed light on
the hypothesised close relations between quantified statements and
conditional probability assertions in the literature \cite<see,
e.g.>{cohen12,2016:SMPS2,PSsubm}. Recent data of Westerners suggest, that in
 task ASP   73\% of the participants respond that the conclusion
 holds, whereas 88\% of the participants respond that the conclusion
 in task     
ASnP does not hold \cite{pfeifertulkki17}, which corresponds to the normative predictions.

We also investigate the well-known MP and its not logically valid but
probabilistically informative counterpart
NMP.  In a sample of  Western participants \cite{pfeifertulkki17}, 68\% responded correctly,
that the conclusion in task MP holds, and 63\% responded correctly
that the conclusion in task NMP does not hold \cite<see also>{pfeifer07a}.  

Finally, as mentioned above, we investigate one of the paradoxes of the material
conditional. 
Western data on Task Prdx indicates that most people (87\% on the
average) understand that
this argument form is probabilistically non-informative \cite{pfeifer10b,pfeifertulkki17}.

\section{Method}
\subsection{Materials and Design}
We used a $2\times 2$ between-participants design where we crossed
task formulations in terms of indicative conditionals versus
formulations in terms of counterfactuals. To control for position effects, we
used two random orders (generated by \texttt{random.org}). This
resulted in four different task booklets.

Each booklet consisted of a brief introduction, of eight tasks, and of 
questions about the booklets (task
difficulty, whether participants took logic or probability classes and
whether they like maths). Furthermore, we included usual demographic
questions at the end. The logical forms of the eight tasks are
explained in Table~\ref{TAB:tasks}. We instantiated these logical forms
into a cover story which was already used in studies on Western samples
\cite<see, e.g.,>{pfeifer10b,pfeifertulkki17}. We adapted and
translated this cover story for the Japanese sample.

For each task, the participants were asked to imagine the following
situation:
\begin{quote}\em
Hanako works in a factory that produces toy blocks. She is
responsible for controlling the production. Every toy block has a
shape (\emph{cylinder, cube \emph{or} pyramid}) and a colour
(\emph{red, blue \emph{or} green}). For example:
\begin{itemize}
\item Red cylinder, red cube, red pyramid
\item Blue cylinder, blue cube, \dots
\item Green cylinder, \dots
\end{itemize}
\end{quote}
Then, for example in task AT1 (indicative conditional), the participants were asked to consider the
following sentence:
\begin{center}
\fbox{\em
  \parbox{.43\textwidth}{\textbf{It is not the case}, that: \,   \textbf{If} the toy
  block is \textbf{not}  a
  \emph{cube}, \textbf{then} the toy block is a
  \emph{cube}.}}
\end{center}
The instructions continued by the following questions, which prompt
answers 
in a forced choice format:
\begin{quote}\em
\noindent Can Hanako infer at all \underline{how sure she can be} that
the sentence in the box holds? \emph{(please tick the appropriate box)}
\begin{itemize}
\item[$\Box$] NO, Hanako can {\bf not} infer how sure she can be that
  the sentence in the box holds.
\item[$\Box$] YES, Hanako can infer how sure she can be that
  the sentence in the box holds. \\
\begin{quote}
\emph{If you chose ``\emph{YES}'', please tick one of the following answers:}
\begin{itemize}
\item[$\Box$] Hanako can be sure that the sentence in the box holds.
\item[$\Box$] Hanako can be sure that the sentence in the box does {\bf
    not} hold.
\end{itemize}
\end{quote}
\end{itemize}
\end{quote}
After each target task, the participants were instructed  to rate  on a scale their subjective confidence in their response. 
The corresponding AT1 task involving counterfactuals was formulated in
exactly the same way with the difference, that the indicative
conditional was replaced by a corresponding counterfactual, as follows:
\begin{center}
\fbox{\em
  \parbox{.43\textwidth}{\textbf{It is not the case}, that: \,   \textbf{If} the toy
  block were \textbf{not}  a
  \emph{cube}, \textbf{then} the toy block would be a
  \emph{cube}.}}
\end{center}
For those tasks involving explicit premises (i.e., in tasks EIn, EI, MP, NMP, and Prdx), we formulated
uncertainties   in terms of verbal descriptions (``\emph{quite sure}''). For instance, consider  task MP:
\begin{quote}\em
\begin{itemize} 
\item[(A)] \dots {\em quite sure} that the toy block is a  \emph{cube}.
\item[(B)] \dots {\em quite sure} that \textbf{if} the toy block is a
  \emph{cube}, \textbf{then} it is \emph{red}.
\end{itemize}
\end{quote}

\subsection{Participants and procedure}

63 Osaka City University undergraduate students 
participated in this study (mean age 20.02 ($SD=1.05$) years, 34
females, 21 males, 8 did not disclose their gender). Their major subjects
included various humanistic fields (3 commerce,   5 culture,  1 geography,  5  history,
4 Japanese, 8        law, 5  linguistics, 1 pedagogy, 2 philosophy, 17
psychology, 2  sociology, and 10 other). Nobody had ever taken logic classes but two
participants had previously taken some probability classes. At the end of the experiment,
participants evaluated the set of tasks as rather difficult (mean 2.76 ($SD=2.11$) on a
scale ranging from 0 (``very difficult'') to 10 (``very
easy'')). 82.54\% reported that they do not like maths.
 
All participants were tested  at the same time during a lesson in a course on cultural psychology.  For reducing the probability for copy-pasting  responses, the
booklets were distributed such that the two task orders and the
two formulations of the conditionals (indicative vs.\ counterfactual)
alternated systematically. Moreover, the experimenter announced that the task booklets differ before the participants started with filling in their responses. The booklets
were formulated in Japanese, the participants' mother tongue.

\section{Results and discussion}

We performed Fisher's exact tests to compare the response  frequencies
among the four booklets (task order 1~$\times$ task order~2 $\times$
indicative conditionals $\times$ counterfactuals) and did not observe
any significant differences after performing Holm-Bonferroni corrections
for multiple significance tests. Likewise, analyses of variance on the
participant's confidence ratings in the correctness of their responses
did not show statistically significant differences among the four
booklets. This replicates previous findings in studies which used Western samples. Specifically, studies on  probabilistic truth
table tasks \cite{over07b,pfeifer15a}  and on  uncertain
argument forms \cite{pfeifertulkki17} did not detect significant difference between indicative conditionals and counterfactuals. Thus, our data speak against
cross-cultural differences between Easterners and Westerners. This calls for further experiments to clarify whether this interesting negative result is due to a too high dissimilarity of our tasks compared to those in other studies on cross-cultural differences. Or, alternatively, whether cross-cultural differences  are not that  strong as they are claimed to be \cite<see, e.g.,>{zhang15}. 

Since there were no significant differences in the responses among the
four booklets, we  pooled the data for the following data analysis
($N=63$). 
Concerning the interpretation of conditionals, we observed high endorsement rates of the conditional probability
hypothesis (see Table~\ref{TAB:percentages}). This is strong support for the hypothesis that  both indicative conditionals and counterfactuals are best modeled by conditional probability.

\begin{table}[!ht]
\begin{center} 
\caption{\renewcommand\baselinestretch{1} 
 Percentages ($n=63$) of ``holds'' (hld), ``does not hold''
 ($\neg$hld), and probabilistic non-informativeness responses (n-inf;
 see also Table~\ref{TAB:tasks}). Predictions based on the conditional
 probability hypothesis of conditionals are in {\bf bold}. Alternative
 hypotheses are indicated in parentheses: $\neg_\supset$ (resp.,
 $\supset_\neg$) denotes wide (resp., narrow) scope negation of the
 material conditional $\supset$; $\wedge$ denotes conjunction. If not
 specified otherwise, predictions coincide.} 
\label{TAB:percentages} 
\vskip 0.12in
\begin{tabular}{rlllll}\hline
             & AT1  &  AT2  &NR     & EIn \\\cline{2-5}

hld:       &{\bf 65.08}$\left(^{\supset_\neg}_\wedge\right)$  & {\bf 76.19}$\left(^{\supset_\neg}_\wedge\right)$ &~6.35  & ~6.45    \rule{0pt}{1.1\normalbaselineskip} \\
\!\!$\neg$hld: &15.87   & 11.11 & {\bf 63.49}$(^{\neg_\supset})$  &{\bf 69.35}      \\
 n-i:    &19.05$(^{\neg_\supset})$& 12.70$(^{\neg_\supset})$ &30.16$\left(^{\supset_\neg}_\wedge\right)$ &24.20 \\[0.1in]
 
             & EI   &MP     &NMP   &  Prdx          \\ \cline{2-5}
 hld:      &{\bf 88.89}  &{\bf 53.97}   & ~9.52  & ~0.00$(^\supset)$\\
\!\!$\neg$hld: &~6.35  & ~3.17   &  {\bf 52.38}& 17.46$(_\wedge)$\\
 n-inf:    & ~4.76 & 42.86   & 38.10&  {\bf 82.54} \\\hline
\end{tabular}
\end{center}
\end{table}

Table~\ref{FIG:conf} presents the mean confidence ratings, which shows how sure the
participants are that their responses are correct. The confidences are relatively
high, with an average value of 7.2 on a rating scale from 0 to 10.

\begin{table}[!ht]
\begin{center} 
\caption{Mean ($M$) and standard deviations ($SD$) of the
  participants' confidence ratings ($n=63$) on a scale from 0 (``very
  sure that my response is not correct'') to 10 (``very sure that my response is correct'';  see also Table~\ref{TAB:tasks}).} 
\label{FIG:conf} 
\vskip 0.12in
\begin{tabular}{rrrrrrrrr}\hline
             & \!\!AT1 &  \!AT2 &\!NR    & \!EIn  & \!EI~   &\!MP    &\!NMP & Prdx  \\ \cline{2-9}
\!\!$M$&\!\!6.77 &\!6.86 &\!7.20 &\!7.71 &\!8.02 &\!7.18 &\!7.02&\!6.82  \\
\!\!$SD$&\!\!1.99 &\!2.06 &\!2.37 &\!1.99 &\!1.97 &\!2.10 &\!2.08&\!1.93  \\\hline

\end{tabular}
\end{center}
\end{table}

\section{Concluding remarks}

Our data suggest that people form their  degree of belief in the
counterfactual \emph{If $A$ were the case, $C$ would be the case} by
equating it with  the corresponding conditional probability of
$C|A$. This is consistent with the observation in previous
experimental work (with Western participants)  that people
treat the factual statement as irrelevant when they  form
their degree of belief in a counterfactual
\cite{pfeifer15a,pfeifertulkki17}. This  can be justified and 
explained by the coherence-based theory of nested conditionals
\cite{GiSa13c,gilio14,2016:SMPS1,GOPSsubm}. Given three events
$A,B,C$ with incompatible $A$ and $B$ (i.e., $A \wedge B$ is a logical
contradiction) the prevision of the conditional random quantity
$((C|B)|A)$ is equal to $p(C|B)$ \cite[Example 1,
p. 225]{GiSa13c}. Thus, the counterfactual \emph{If $A$ were the case,
  $C$ would be the case} can be modeled by the degree of belief in the
conditional random quantity $(C|A)|\neg A$ which equals to $p(C|A)$
(i.e., $Prevision((C|A)|\neg A)=p(C|A)$). This is an explanation for
why people---as experimentally demonstrated in Western samples and
also in our Japanese sample---respond by corresponding conditional
probabilities when asked to give a degree of belief in a 
counterfactual.

 Our data suggest a negative answer to the question whether there
are cross-cultural differences between Easterners and Westerners
w.~r.~t.\ reasoning about indicative conditionals, counterfactuals, and
their negations. Further experimental work is needed to 
substantiate the hypothesis that conditional probability is the
\emph{universal} key ingredient for psychological theories of  
indicative conditionals and counterfactuals. 

Finally, we note that  
adaptation of reasoning styles can be one of the universal adaptive strategies    across cultures. 
The question of which aspects of human reasoning are universal, and in how far  they are universal, is important and calls for collaborations of psychologists of reasoning and cultural psychologists.


\section{Acknowledgments}

Niki Pfeifer is supported by the DFG project
    PF~740/2-2 as part of the Priority Program ``New Frameworks of Rationality'' (SPP1516).


\end{document}